\documentclass[runningheads]{llncs}
\usepackage{xfrac, url}
\usepackage{amsmath}
\usepackage{amssymb}
\usepackage{tikz}
\usepackage{footnote}
\usepackage{rotfloat}
\usetikzlibrary{matrix}
\usepackage[table]{colortbl}

\title{11 $\times$ 11 Domineering is Solved: \\ The first player wins}
\author{Jos W.H.M. Uiterwijk}
\institute{Department of Data Science and  Knowledge Engineering (DKE) \\ 
Maastricht University, Maastricht, The Netherlands
\email{uiterwijk@maastrichtuniversity.nl}}

\begin{document}

\maketitle

\begin{abstract}
We have developed a program called {\sc MUDoS} (Maastricht University Domineering Solver) that solves Domineering positions in a very efficient way. This enables the solution of known positions so far (up to the $10 \times 10$ board) much quicker (measured in number of investigated nodes). 

More importantly, it enables the solution of the $11 \times 11$ Domineering board, a board up till now far out of reach of previous Domineering solvers.
The solution needed the investigation of 259,689,994,008 nodes, using almost half a year of computation time on a single simple desktop computer. The results show that under optimal play the first player wins the $11 \times 11$ Domineering game, irrespective if Vertical  or Horizontal  starts the game.

In addition, several other boards hitherto unsolved were solved. Using the convention that Vertical starts, the $8 \times 15$, $11 \times 9$,  $12 \times 8$, $12 \times 15$,  $14 \times 8$, and $17 \times 6$ boards are all won by Vertical, whereas the $6 \times 17$,  $8 \times 12$, $9 \times 11$, and  $11 \times 10$  boards are all won by Horizontal. 
\end{abstract}

%======================================================================

\section{Introduction}
Domineering is a two-player perfect-information game invented by G{\"o}ran Andersson around 1973. It was popularized to the general public in an article by Martin Gardner \cite{Gardner1974}. It can be played on any subset of a square lattice, though mostly it is restricted to rectangular $m \times n$ boards, where $m$ denotes the number of rows and $n$ the number of columns. The version introduced by Andersson and Gardner was the $8 \times 8$ board.

Play consists of the two players alternately placing a $1 \times 2$ tile (domino) on the board, where the first player may place the tile only in a vertical alignment, the second player only horizontally. The first player being unable to move loses the game, his opponent (who made the last move) being declared the winner. Since the board is gradually filled, i.e., Domineering is a converging game, the game always ends, and ties are impossible. With these rules the game belongs to the category of {\em combinatorial games}, for which a whole theory (the Combinatorial Game Theory, or CGT in short) has been developed.

Among combinatorial game theorists Domineering received quite some attention, but this was limited to rather small or irregular boards \cite{ANW2007,Berlekamp1988,BCG1982,Conway1976,Kim1996,Wolfe1993}. Larger (rectangular) boards were solved using $\alpha$-$\beta$ search \cite{KM1975}, leading to solving all boards up to the standard $8 \times 8$ board \cite{BUvdH2000}, later extended to the $9 \times 9$ board \cite{vdHUvR2002,UvdH2000}, and finally extended to larger boards up to $10 \times 10$ \cite{Bullock2002a,Bullock2002b}.

%======================================================================

\section{Three Approaches}

The following subsections give a rough characterization of the two main programs used to systematically solve Domineering positions so far, and of the program used to obtain the new results, as described in this paper.

%========================

\subsection{A Brute-Force Appoach: {\sc Domi}}
%Although small rectangular boards up to the $6 \times 6$ board (and some wider (higher)  boards for low height (width)) were solved already by combinatorial game theory researchers \cite{BCG1982,Conway1976}, the 
The first systematic analysis of rectangular Domineering boards was performed by Breuker et al. \cite{BUvdH2000,vdHUvR2002,UvdH2000}. They developed the program {\sc Domi}, using a straightforward variant of the $\alpha$-$\beta$ technique \cite{KM1975}, enhanced with a transposition table. The algorithm did not use perfect domain knowledge for classifying positions as wins or losses and hence can be characterized as a pure brute-force approach. Transposition tables with 2M ($2^{21}$) entries were used with a two-level replacement scheme called {\em TwoBig} \cite{BUvdH1996}, in which each entry can store two table positions. Mirror symmetries are taken into account. The newest position is always stored, overwriting the less important position in terms of nodes investigated.

%========================

\subsection{A Knowledge-Based Approach: {\sc Obsequi}}
A few years later Nathan Bullock published results on solving Domineering boards up to the $10 \times 10$ board \cite{Bullock2002b}. His program {\sc Obsequi} used a sophisticated evaluation function which can determine statically the winner at a shallower point in the search tree than {\sc Domi} did. This allowed the elimination of large portions of the search space, resulting in much more efficient solving of Domineering boards. {\sc Obsequi} used a transposition table (taking mirror symmetries into account) with $2^{23}$  entries with either a two-level {\em TwoBig} replacement scheme or a one-level replacement scheme called {\em FindFirst} \cite{Bullock2002a}. Also, a much better move-ordering heuristic was used, plus the use of a dominance relation to prune provably irrelevant moves. Since the main advantage of Bullock's program is based on game-specific knowledge, we can characterize his approach as a knowledge-based approach.

%========================

\subsection{A Knowledge-Intensive Approach: {\sc MUDoS}}
Uiterwijk continued using game-specific knowledge to an even more detailed extent. His program {\sc MUDoS} incorporated deep knowledge of Domineering positions with known result. These knowledge rules are so intense, that it even enables solving many game boards without any search at all (i.e., investigating a single node, the empty board under consideration). This was called {\em perfectly solving} \cite{Uiterwijk2014a}. The most important feature of these knowledge rules is the number of safe moves that a player provably can reach in a  position \cite{Uiterwijk2014b,Uiterwijk2014c,Uiterwijk2014d}. The transposition table used (again taking mirror symmetries into account) contained $2^{26}$ entries, with a simple one-level {\em Deep} replacement scheme.
Due to the heavy use of very knowledge-intense rules based on game-specific properties we can characterize his approach as a knowledge-intensive approach.

%======================================================================

\section{New Results}

After almost half a year of computation time, $11 \times 11$ Domineering was solved. We give some data in Section \ref{sub:11x11}. As a sidetrack, we  solved several other new boards. Data are given in Section \ref{sub:others}. An overview of updated combinatorial-game-theoretic values of Domineering boards is given in Section \ref{sub:overview}.

\subsection{The Solution of $11 \times 11$ Domineering{\label{sub:11x11}}}
The solution of $11 \times 11$ Domineering took 174 days and 15 hours on a standard desktop computer (a HP with duo core Intel E8400 3.00 GHz CPU with a 64-bit Windows 7 operating system and 4 GB internal memory). The {\sc MUDoS} program is written in C\#.

The result is that the first player under optimal play wins the game. Since the board is square, this is irrespective of Vertical or Horizontal moving first.

To put the solution of the $11 \times 11$ board  into perspective, we show in Table \ref{tab:11x11} the results and number of nodes investigated to solve square boards up to $11 \times 11$ by the three programs mentioned in the previous section.

\begin{savenotes}
\begin{table}[h]
\caption{Results and number of nodes investigated to solve square Domineering boards. Vertical always starts. A ``1'' and ``2'' in the results column indicate a first-player (Vertical) and second-player (Horizontal) win, respectively. A ``--'' in a column indicates that the program was unable to solve the position.} 
\label{tab:11x11}
\centering
\begin{tabular}{| c | c | r | r |  r |}
\hline
board size  		& result & {\sc Domi} \cite{BUvdH2000}	& {\sc Obsequi} \cite{Bullock2002b}		& {\sc MUDoS}		\\
\hline
$2 \times 2$		& 1	& 1			&  1				& 1				\\
$3 \times 3$		& 1	& 1 			&  1				& 1				\\
$4 \times 4$		& 1	&  40			&  23				& 1				\\
$5 \times 5$		& 2	&  604			&  259				& 17				\\
$6 \times 6$		& 1	&  17,232		&  908				& 1				\\
$7 \times 7$		& 1	&  408,260		&  31,440			& 1				\\
$8 \times 8$		& 1	& 441,990,070	&  2,023,301			& 24,147			\\
$9 \times 9$\		& 1	&  $\sim$25,000,000,000\footnote{This result was obtained with an improved version of {\sc Domi}, around 2000 \cite{Breuker2014}. The exact number of nodes investigated was lost.}			
							&  1,657,032,906		& 4,917,736			\\
$10 \times 10$	& 1	&  --			&  3,541,685,253,370	& 13,506,805		\\
$11 \times 11$	& 1	&  --			&  --				& 259,689,994,008		\\
\hline
\end{tabular}
\end{table}
\end{savenotes}

For the result the investigation of 259,689,994,008 nodes was needed, with an average speed of 17,211 nodes/sec. While this is some ten times slower than {\sc Obsequi}'s speed, this decrease in speed is by far compensated by the much higher pruning efficiency, as evidenced by the ratio's of the number of nodes investigated by {\sc MUDoS} and {\sc Obsequi}. For the $8\times 8$, $9 \times 9$ and $10 \times 10$ boards these are 1.19\%, 0.30\%, and 0.00038\%, respectively. Of course the latest number is so low, since {\sc Obsequi} solved the $10 \times 10$ board on a distributed network of several computers (no further details given), without memory sharing, by which transposition tables will be far less effective. But as a striking fact, whereas {\sc Obsequi} needed several months of computation time on this network, {\sc MUDoS} needs only 21 minutes on a single computer to solve the $10 \times 10$ board.

%========================

\subsection{The Solution of New Other Domineering Boards{\label{sub:others}}}
Besides $11 \times 11$ Domineering we were able to solve several other new Domineering boards. The results are given in Table \ref{tab:other}.

\begin{table}[h]
\caption{Results and number of nodes investigated to solve other new Domineering boards. Vertical always starts. A ``1'' and ``2'' in the results column indicate a first-player (Vertical) and second-player (Horizontal) win, respectively. } 
\label{tab:other}
\centering
\begin{tabular}{| c | c | r || c | c | r |}
\hline
board size  		& result 	& \# nodes		& board size		& result 	& \# nodes		\\
\hline
			& 		& 			& $11 \times 10$	& 2		& 1			\\
$9 \times 11$	& 2	& 84,145,153	&  $11 \times 9$		& 1		& 23,183,077	\\
$6 \times 17$	& 2 	& 25,670,138,842	&  $17 \times 6$		& 1		& 810,774,495	\\
$8 \times 12$	& 2	& 273,559,795	&  $12 \times 8$		& 1		& 11,960,354	\\
			& 	& 			&  $14 \times 8$		& 1		& 490,146,677	\\
$8 \times 15$	& 1	& 1			& 				&		&			\\
$12 \times 15$	& 1	& 1			&				&		&			\\
\hline
\end{tabular}
\end{table}

The most notable results and their consequences are given below. We there use the notion of outcome class \cite{Conway1976,BCG1982,ANW2007} of an $m \times n$ board, denoted by $[m \times n]$, where an outcome class is N, P, V, or H (1st = Next player; 2nd = Previous player; Vertical, irrespective of who starts;  Horizontal, irrespective of who starts).

\vspace*{-0.2cm}
\subsubsection*{Other boards with width or heigth 11 \\}
Although the $10 \times 11$ board was already solved (Vertical wins), using the translational symmetry rules of Lachmann c.s. \cite{LMR2002}, and even perfectly solved \cite{Uiterwijk2014a}, the $11 \times 10$ board was not. However, MUDoS solves it investigating just 1 node, showing that Horizontal wins.\footnote{We note that solving a board investigating a single node is not exactly the same as perfectly solving a board, since in the latter the board is solved using characteristics of the board solely, without generating the possible moves, whereas in the former the possible moves are generated, but immediately proven to contain at least one winning move or only losing moves.} As a result  $[10 \times 11]$ = V (and $[11 \times 10]$ = H).
Further, with some more work, we were able to solve the $9 \times 11$ board (Horizontal wins) and the $11 \times 9$ board (Vertical wins).  Consequently,  $[9 \times 11]$ = H (and $[11 \times 9]$ = V).

\vspace*{-0.2cm}
\subsubsection*{Boards with width or heigth 6 \\}
The $6 \times 17$ and $17 \times 6$ boards were also solved (wins for Horizontal and Vertical, respectively). Consequently, $[6 \times 17]$ = H. Moreover, using the translational symmetry rules of Lachmann c.s. \cite{LMR2002} and the facts that $[6 \times 4]$ = N and $[6 \times n]$ with $n = 8$, 12, and 14 are H, it follows that $[6 \times 21]$ (17+4) = N or H, and $[6 \times 25]$ (17+8) = H, $[6 \times 29]$ (17+12) = H,  and $[6 \times 31]$ (17+14) = H. Moreover, in \cite{DrummondCole2014} it was shown that $[6 \times n]$ for $n > 31$ = N or H for widths 33, 35, 37, 39, 43, 45, 47, 51, and 59. Using the result for $[6 \times 17]$ all these values analogously are determined to be H, the only exception being width 35 (still N or H). This shows that the holes in the results for boards of height 6 have considerably been filled. The outcome classes for all $6 \times n$ boards are known now, the only exceptions being the  $6 \times 18$, $6 \times 21$, $6 \times 23$, $6 \times 27$, and $6 \times 35$ boards, all five having outcome classes N or H, which means that Horizontal at least wins as first player. Of course the results for $[m \times 6]$ can similarly be updated, replacing H by V.

\vspace*{-0.2cm}
\subsubsection*{Boards with width or heigth 8 \\}
The $8 \times 12$ and $12 \times 8$ boards were also solved (wins for Horizontal and Vertical, respectively). Consequently, $[8 \times 12]$ = H, but also, using the translational symmetry rules and the facts that $[8 \times 10]$ and $[8 \times 16]$ are H, it follows that $[8 \times 22]$ (12+10) = H, $[8 \times 24]$ (12+12) = H, and $[8 \times 28]$ (12+16) = H. Moreover, since $[8 \times 10]$ = H and all $[8 \times n]$ for even $n$ from 20--28 are H, it follows that all $[8 \times n]$ with even $n \ge 20$ are H. This makes the entries in the $8 \times n$ row completely regular for even $n$ from $n = 20$ onwards, in contrast to \cite{DrummondCole2014}, were (in an irregular way) some of those were determined to be H, the others as N or H.
We also were able to solve the $14 \times 8$ board (Vertical wins), but not the $8 \times 14$ board yet. This means that $[8 \times 14]$ = N or H.  This leaves the $8 \times 14$ and $8 \times 18$ boards as the only holes in this row for even width. Finally, the $8 \times 15$ (and $12 \times 15$) board is trivially solved to be a Vertical win (so outcome class N or V), but the rotated $15 \times 8$ (and $15 \times 12$) board could not yet be determined. Again, of course the results for $[m \times 8]$ can similarly be updated, replacing H by V, including that all $[m \times 8]$ with even $m \ge 20$ are V.

%========================

\subsection{Updated Table of CGT Values of Domineering \label{sub:overview}}

In Table \ref{tab:complete} we give a complete updated overview of all results for solved Domineering boards,  as outcome classes. The results are taken from \cite{DrummondCole2014} and includes results from \cite{Conway1976,BCG1982,Berlekamp1988,BUvdH2000,UvdH2000,Bullock2002b,vdHUvR2002,LMR2002,DrummondCole2014}.\footnote{Although Drummond-Cole determined the outcome classes for $8 \times 26$ and $26 \times 8$ (H and V), these results were not included in his table of known outcome classes for Domineering \cite{DrummondCole2014}.} 
In addition, our new results have been added. This table is also available at \cite{Uiterwijk2015}, where any future updates will be made public.

%\newpage

\setlength{\tabcolsep}{2pt}
\begin{sidewaystable}[H]
%\begin{table}[h!]
\caption{Updated results for outcome classes of Domineering boards. An entry like NH means that the value is either N or H. -V (or -H) means that all we know is that the outcome class is not V (or H). The notes are explained in the text. New results obtained are shaded.} 
\label{tab:complete}
%\centering
\vspace*{0.3cm}
\hspace*{-0.3cm}
\begin{tabular}{| c || c | c | c | c | c | c | c | c | c | c | c | c | c | c | c | c | c | c | c | c | c | c | c | c | c | c | c | c | c | c | c | c | c | c | c |}
\hline
%$m$$\backslash$$n$  	& 0 & 0 & 0 & 0 & 0 & 0 & 0 & 0 & 0 & 1 & 1 & 1 & 1 & 1 & 1 & 1 & 1 & 1 & 1 & 2 & 2 & 2 & 2 & 2 & 2 & 2 & 2 & 2 & 2 & 3 & 3 & $>$	\\
$m$$\backslash$$n$  	& 1 & 2 & 3 & 4 & 5 & 6 & 7 & 8 & 9 & 10 & 11 & 12 & 13 & 14 & 15 & 16 & 17 & 18 & 19 & 20 & 21 & 22 & 23 & 24 & 25 & 26 & 27 & 28 & 29 & 30 & 31 & $>31$	\\
\hline
\hline
1 & P & H & H & H & H & H & H & H & H & H & H & H & H & H & H & H & H & H & H & H & H & H & H & H & H & H & H & H & H & H & H & H	\\
\hline
2 & V & N & N & H & V & N & N & H & V & N & N & H & P & N & N & H & H & N & N & H & H & H & N & H & H & H & 1 & H & H & H & H & H	\\
\hline
3 & V & N & N & H & H & H & H & H & H & H & H & H & H & H & H & H & H & H & H & H & H & H & H & H & H & H & H & H & H & H & H & H	\\
\hline
4 & V & V & V & N & V & N & V & H & V & H & V & H & P & H & H & H & H & H & H & H & H & H & H & H & H & H & H & H & H & H & H & H		\\
\hline
5 & V & H & V & H & P & H & H & H & H & H & H & H & H & H & H & H & H & H & H & H & H & H & H & H & H & H & H & H & H & H & H & H	\\
\hline
6 & V & N & V & N & V & N & V & H & V & N & N & H & V & H & N & H & \cellcolor{gray}H & NH & N & H & \cellcolor{gray}NH & H & NH & H & \cellcolor{gray}H & H & NH & H & \cellcolor{gray}H & H & \cellcolor{gray}H & 1)		\\
\hline
7 & V & N & V & H & V & H & N & H & H & H & H & H & H & H & H & H & H & H & H & H & H & H & H & H & H & H & H & H & H & H & H & H	\\
\hline
8 & V & V & V & V & V & V & V & N & V & H & V & \cellcolor{gray}H & V & \cellcolor{gray}NH & \cellcolor{gray}NV & H &  & NH &  & H &  & \cellcolor{gray}H &  & \cellcolor{gray}H &  & H &  & \cellcolor{gray}H &  & H &  & 2)	\\
\hline
9 & V & H & V & H & V & H & V & H & N & H & \cellcolor{gray}H & H & H & H & H & H & H & H & H & H & H & H & H & H & H & H & H & H & H & H & H & H	\\
\hline
10 & V & N & V & V & V & N & V & V & V & N & \cellcolor{gray}V &  & V &  & NV &  &  &  &  & H &  & NH &  &  &  & NH &  &  &  & NH &  & 	\\
\hline
11 & V & N & V & H & V & N & V & H & \cellcolor{gray}V & \cellcolor{gray}H & \cellcolor{gray}N & H & --V & H & NH & H & NH & H & NH & H & NH & H & NH & H & NH & H & NH & H & NH & H & NH & H	\\
\hline
12 & V & V & V & V & V & V & V & \cellcolor{gray}V & V &  & V & NP & V &  & \cellcolor{gray}NV &  &  &  &  &  &  &  &  & H &  &  &  &  &  &  &  & 	\\
\hline
13 & V & P & V & P & V & H & V & H & V & H & --H & H & NP & H & --V & H & NH & H & NH & H & NH & H & NH & H & NH & H & NH & H & NH & H & NH & 3)	\\
\hline
14 & V & N & V & V & V & V & V & \cellcolor{gray}NV & V &  & V &  & V & NP & NV &  &  &  &  &  &  &  &  &  &  &  &  & H &  & NH &  & 	\\
\hline
15 & V & N & V & V & V & N & V & \cellcolor{gray}NH & V & NH & NV & \cellcolor{gray}NH & --H & NH & NP &  &  & NH &  &  &  &  &  &  &  &  &  &  &  & H &  & 	\\
\hline
16 & V & V & V & V & V & V & V & V & V &  & V &  & V &  &  & NP &  &  &  &  &  &  &  &  &  &  &  &  &  &  &  & 	\\
\hline
17 & V & V & V & V & V & \cellcolor{gray}V & V &  & V &  & NV &  & --H &  &  &  & NP &  &  &  &  &  &  &  &  &  &  &  &  &  &  & 	\\
\hline
18 & V & N & V & V & V & NV & V & NV & V &  & V &  & V &  & NV &  &  & NP &  &  &  &  &  &  &  &  &  &  &  &  &  & 	\\
\hline
19 & V & N & V & V & V & N & V &  & V &  & NV &  & NV &  &  &  &  &  & NP &  &  &  &  &  &  &  &  &  &  &  &  & 	\\
\hline
20 & V & V & V & V & V & V & V & V & V & V & V &  & V &  &  &  &  &  &  & NP &  &  &  &  &  &  &  &  &  &  &  & 	\\
\hline
21 & V & V & V & V & V & \cellcolor{gray}NV & V &  & V &  & NV &  & NV &  &  &  &  &  &  &  & NP &  &  &  &  &  &  &  &  &  &  & 	\\
\hline
22 & V & V & V & V & V & V & V & \cellcolor{gray}V & V & NV & V &  & V &  &  &  &  &  &  &  &  & NP &  &  &  &  &  &  &  &  &  & 	\\
\hline
23 & V & N & V & V & V & NV & V &  & V &  & NV &  & NV &  &  &  &  &  &  &  &  &  & NP &  &  &  &  &  &  &  &  & 	\\
\hline
24 & V & V & V & V & V & V & V & \cellcolor{gray}V & V &  & V & V & V &  &  &  &  &  &  &  &  &  &  & NP &  &  &  &  &  &  &  & 	\\
\hline
25 & V & V & V & V & V & \cellcolor{gray}V & V &  & V &  & NV &  & NV &  &  &  &  &  &  &  &  &  &  &  & NP &  &  &  &  &  &  & 	\\
\hline
26 & V & V & V & V & V & V & V & V & V & NV & V &  & V &  &  &  &  &  &  &  &  &  &  &  &  & NP &  &  &  &  &  & 	\\
\hline
27 & V & N & V & V & V & NV & V &  & V &  & NV &  & NV &  &  &  &  &  &  &  &  &  &  &  &  &  & NP &  &  &  &  & 	\\
\hline
28 & V & V & V & V & V & V & V & \cellcolor{gray}V & V &  & V &  & V & V &  &  &  &  &  &  &  &  &  &  &  &  &  & NP &  &  &  & 	\\
\hline
29 & V & V & V & V & V & \cellcolor{gray}V & V &  & V &  & NV &  & NV &  &  &  &  &  &  &  &  &  &  &  &  &  &  &  & NP &  &  & 	\\
\hline
30 & V & V & V & V & V & V & V & V & V & NV & V &  & V & NV & V &  &  &  &  &  &  &  &  &  &  &  &  &  &  & NP &  & 	\\
\hline
31 & V & V & V & V & V & \cellcolor{gray}V & V &  & V &  & NV &  & NV &  &  &  &  &  &  &  &  &  &  &  &  &  &  &  &  &  & NP & 	\\
\hline
$>$ 31 & V & V & V & V & V & 4) & V & 5) & V &  & V &  & 6) &  &  &  &  &  &  &  &  &  &  &  &  &  &  &  &  &  &  & 	\\
\hline
\end{tabular}
%\end{table}
\end{sidewaystable}

%\vspace*{-0.5cm}
In this table the following notes apply: 
1) the outcome classes for all $n > 31$ are H, except that the outcome class for $n = 35$ is N or H; 
2) the outcome classes for all even $n \ge 20$ are H; 
3) the outcome classes are alternating H (even $n$) and N or H (odd $n$); 
4)-6): equivalent to notes 1)-3) by replacing $n$ with $m$ and H with V.

For boards with one or both dimensions larger than 31, besides the results in the notes above, nothing is known about their outcome classes, except of course that $m \times m$ boards have outcome classes N or P, that $m \times 2km$ boards have outcome classes H, and that $2kn \times n$ boards have outcome classes V.

%======================================================================

\section{Conclusions and Future Work}
As can be seen from the results it is clear that {\sc MUDoS} is a very efficient Domineering solver. All boards solved before are solved in an equal amount (for the trivial boards) or far smaller (for the more complex boards) number of investigated nodes than by previous solvers.

The efficiency of our solver enabled the solution of the $11 \times 11$ Domineering board. The result indicates that the first player wins. Moreover, several new rather complex boards have been solved. Applying these together with the use of the translational symmetry rules  updated the Domineering  outcome class landscape considerably.

Regarding future work,  foremost this condensed overview will be extended to a full publication. This will include a detailed description of {\sc MUDoS}' knowledge rules and heuristics employed. Moreover, the impact of the rules and heuristics on solving performance, separately and in combination, will be illlustrated with experiments.

As a follow-up we moreover intend as a last step to enhance the solving power of our Domineering program by incorporating knowledge from Combinatorial Game Theory into our solver. A preliminary experiment using endgame databases up to 16 squares filled with CGT values, combined with a very simplistic $\alpha$-$\beta$ solver showed reductions up to 99\% for boards up to $7 \times 7$ \cite{BU2014}.

%======================================================================

%\newpage


\begin{thebibliography}{[MT1]}

\bibitem{ANW2007}
Albert, M.H., Nowakowski, R.J., and Wolfe, D.:
{\it Lessons in Play: An Introduction to Combinatorial Game Theory}.
A K Peters, Wellesley, MA (2007)

\bibitem{BU2014}
Barton, M., and  Uiterwijk, J.W.H.M.:
Combining Combinatorial Game Theory with an $\alpha$-$\beta$ Solver for {D}omineering,
In: {\it BNAIC 2014 - Proceedings of the 26th Benelux Conference on Artificial Intelligence} 
(Eds. F. Grootjen, M. Otworowska, and J. Kwisthout), Radboud University, Nijmegen (2014) 9--16

\bibitem{Berlekamp1988}
Berlekamp, E.R.:
Blockbusting and {D}omineering.
{\it J. Combin. Theory (Ser.~A)} {\bf 49} (1988) 67--116

\bibitem{BCG1982}
Berlekamp, E.R., Conway, J.H., and Guy, R.K.:
{\it Winning Ways for your Mathematical Plays}, volumes 1--2.
Academic Press, London  (1982);
2nd edition, in four volumes: vol. 1 (2001), vols. 2, 3 (2003), vol. 4 (2004). A K Peters, Wellesley, MA

\bibitem{Breuker2014}
Breuker, D.M., personal communication (2014)

\bibitem{BUvdH1996}
Breuker, D.M., Uiterwijk, J.W.H.M., and Herik, H.J. van den:
Replacement schemes and two-level tables.
{\it ICCA Journal} {\bf 19} (1996) 175--180

\bibitem{BUvdH2000}
Breuker, D.M., Uiterwijk, J.W.H.M., and Herik, H.J. van den:
Solving $8\times 8$ {D}omineering.
{\it Theoret. Comput. Sci. (Math Games)} {\bf 230} (2000) 195--206

\bibitem{Bullock2002a}
Bullock, N.:
{\it Domineering: Solving Large Combinatorial Search Spaces}.
M.Sc. thesis, University of Alberta (2002) 

\bibitem{Bullock2002b}
Bullock, N.:
Domineering: Solving large combinatorial search spaces.
{\it ICGA Journal} {\bf 25} (2002) 67--84

\bibitem{Conway1976}
Conway, J.H.:
{\it On Numbers and Games}.
Academic Press, London (1976)

\bibitem{DrummondCole2014}
Drummond-Cole, G.C.:
An update on {D}omineering on rectangular boards.
{\it Integers} {\bf 14} (2014) 1--13

\bibitem{Gardner1974}
Gardner, M.: 
Mathematical Games.
{\it Scientific American} {\bf 230} (1974) 106--108

\bibitem{vdHUvR2002}
Herik, H.J. van den, Uiterwijk, J.W.H.M., and Rijswijck, J. van:
Games solved: Now and in the future.
{\it Artificial Intelligence} {\bf 134} (2002) 277--311

\bibitem{Kim1996}
Kim, Y.: 
New values in {D}omineering.
{\it Theoret. Comput. Sci. (Math Games)} {\bf 156} (1996) 263--280

\bibitem{KM1975}
Knuth, D.E. and Moore, R.W.:
An analysis of alpha-beta pruning.
{\it Artificial Intelligence} {\bf 6} (1975) 293--326

\bibitem{LMR2002}
Lachmann, M., Moore, C. and Rapaport, I.:
Who wins {D}omineering on rectangular boards?
In: {\it More Games of No Chance}
(ed. R.J. Nowakowski),
Cambridge University Press, Cambridge;
MSRI Publications {\bfseries 42} (2002) 307--315

\bibitem{UvdH2000}
Uiterwijk, J.W.H.M. and Herik, H.J. van den:
The advantage of the initiative.
{\it Information Sciences} {\bf 122} (2000) 43--58

\bibitem{Uiterwijk2014a}
Uiterwijk, J.W.H.M.:
Perfectly Solving Domineering Games. 
In: {\em Computer games, Workshop on Computer games, CGW at IJCAI 2013, Beijing, China, revised Selected Papers}
(eds. T. Cazenave, M.H.M. Winands, and H. Iida). 
{\it Communications in Computer and Information Science} {\bf 408} (2014) 97--121

\bibitem{Uiterwijk2014b}
Uiterwijk, J.W.H.M.:
The Impact of Safe Moves on Perfectly Solving Domineering Boards. Part 1: Analysis and Experiments with 1-Step Safe Moves. 
{\it ICGA Journal} {\bf 37(2)} (2014) 97--105

\bibitem{Uiterwijk2014c}
Uiterwijk, J.W.H.M.:
The Impact of Safe Moves on Perfectly Solving Domineering Boards. Part 2: Analysis and Experiments with Multi-Step Safe Moves. 
{\it ICGA Journal} {\bf 37(3)} (2014) 144--160

\bibitem{Uiterwijk2014d}
Uiterwijk, J.W.H.M.:
The Impact of Safe Moves on Perfectly Solving Domineering Boards. Part 3: Theorems and Conjectures.
{\it ICGA Journal} {\bf 37(4)} (2014) 207--213

\bibitem{Uiterwijk2015}
Uiterwijk, J.W.H.M.: 
Updated game theoretic values for {d}omineering boards.
\url {https://dke.maastrichtuniversity.nl/jos.uiterwijk/?page_id=39}

\bibitem{Wolfe1993}
Wolfe, D.:
Snakes in {D}omineering games.
{\it Theoret. Comput. Sci. (Math Games)} {\bf 119} (1993) 323--329

\end{thebibliography}
\end{document}